%% file: main.tex
\documentclass{article}

\usepackage{microtype}
\usepackage{graphicx}
\usepackage{subfigure}
\usepackage{amsmath,amssymb} 
\usepackage{empheq, nccmath} 
\usepackage{booktabs} 

\usepackage{natbib} 
\usepackage{tikz} 
\usepackage{multirow} 

\usepackage{hyperref}

\usepackage{amsmath}
\usepackage{amsfonts}
\usepackage{mathtools}
\usepackage[mathscr]{euscript}
\usepackage{algorithm}
\usepackage{algpseudocode}
\usepackage{amsthm}
\usepackage[normalem]{ulem} 

\def\pr{\mathbb{P}}

\def\X{{\mathbf{X}}}
\def\x{{\mathbf{x}}}

\def\A{{\mathcal A}}
\def\HPD{{\text{HPD}}}

\newcommand{\david}[1]{\textbf{\color{teal}[David: #1]}}

\newcommand{\bpl}[1]{{\color{blue}#1}}

\usepackage[accepted]{icml2021}

\icmltitlerunning{Practical Local Conformal Inference in High Dimensions
}

\begin{document}

\twocolumn[
\icmltitle{MD-split+: Practical Local Conformal Inference in High Dimensions 
}



\icmlsetsymbol{equal}{*}

\begin{icmlauthorlist}
\icmlauthor{Benjamin LeRoy}{equal,cmu}
\icmlauthor{David Zhao}{equal,cmu}
\end{icmlauthorlist}

\icmlaffiliation{cmu}{Department of Statistics and Data Science, Carnegie Mellon University, Pittsburgh, Pennsylvania, USA}

\icmlcorrespondingauthor{Benjamin LeRoy}{bpleroy@stat.cmu.edu}

\icmlkeywords{Conformal Inference, Local Conformal Inference, Model Diagnostics, MD-split+, distribution-free inference, ICML}

\vskip 0.3in
]



\printAffiliationsAndNotice{\icmlEqualContribution} 

\begin{abstract}
\input{sections/abstract}

\end{abstract}

\section{Introduction}\label{sec:intro}

\input{sections/intro}



\section{Related Work}\label{sec:related_work}
\input{sections/related_work}



\section{Methods}\label{sec:methods}
\input{sections/methods}

\section{Examples} \label{sec:examples}
\input{sections/examples}

\section{Discussion}\label{sec:discussion}
\input{sections/discussion}

\section{Acknowledgements} \label{sec:ack}
\input{sections/acknowledgements}

\bibliography{BensProposalRelatedPapers}

\end{document}

%% file: sections/abstract.tex

Quantifying uncertainty in model predictions is a common goal for practitioners seeking more than just point predictions. One tool for uncertainty quantification that requires minimal assumptions is conformal inference, which 
can help create probabilistically valid prediction regions for black box models. 
Classical conformal prediction only provides \textit{marginal} validity, whereas in many situations \textit{locally} valid prediction regions are desirable. 
Deciding how best to partition the feature space $\mathcal{X}$ 
when applying localized conformal prediction is still an open question. 
We present \texttt{MD-split+}, a \textit{practical} local conformal approach that creates 
$\mathcal{X}$ partitions 
based on localized model performance of conditional density estimation models. Our method handles complex real-world data settings where such models may be misspecified, and scales to high-dimensional inputs. 
We discuss how our local partitions philosophically align with 
expected behavior from an unattainable conditional conformal inference approach. 
We also empirically compare our method against other local conformal approaches.


%% file: sections/intro.tex

Quantifying uncertainty in model predictions is a 
fundamental goal of predictive inference. 
In this setting, the practitioner is interested in using covariates or features $\X \in \mathcal{X} \subseteq \mathbb{R}^d$ 
to learn about a response variable $Y \in \mathbb{R}$.
Without assumptions on the distribution $F$ of the underlying data, it is challenging to create valid prediction regions for $Y|\X =\x$. One tool for uncertainty quantification that requires minimal distributional assumptions is conformal inference, which can be applied to 
a range of black box models. 
The 
classical version of conformal inference \citep{Vovk2005} creates \textit{marginally valid} prediction regions: \[
\mathbb{P}_{(\X,Y) \sim F}(Y\in C(\x)) > 1 -\alpha\;,
\]
where $C(\x) \subseteq \mathbb{R}$ denotes the prediction region for $Y$ given $\X = \x$. $C(\x)$ provides $1-\alpha$ level \textit{coverage} for $Y$, averaged across the marginal distribution of $\X$.

Marginal validity may not be sufficient in practice when working with large, heterogeneous populations. For example, suppose $Y$ is the effect of a cholesterol-lowering medication on a patient, and $\X$ includes the patient's characteristics. Marginally valid prediction regions might be extremely precise for patients weighing less than 200 lbs., but be correct 0\% of the time for patients weighing more than 200 lbs. Patients in the latter category are not helped by the guarantee that prediction regions on average over all patients achieve $1-\alpha$ level coverage. 

To address this concern, several papers 
\citep{Lei2014, Barber2019a, guan2019localconf} have highlighted the 
desirable properties of \textit{local conformal inference}, which 
provides the stronger guarantee of \textit{locally valid} prediction regions. 
Local validity ensures $1-\alpha$ level coverage not only on average across $\X$, but also within predefined local regions $A$ of the $\mathcal{X}$ space:
\[
\mathbb{P}_{(\X,Y) \sim F}(Y\in C(\x) |\x \in A) > 1 -\alpha\,,\ \forall A \in \mathcal{A},
\] where $\mathcal{A}$ is a partition of $\mathcal{X}$. 
Returning to the cholesterol medication example, a local conformal approach might define bins based on the patient's weight (e.g. 100-120 lbs., 120-140 lbs., etc.), and thus ensure $1-\alpha$ level coverage within each subpopulation. \citet{Lei2014}'s approach is similar in spirit to this, defining regions $A$ as local segments of $\mathcal{X}$ using Euclidean distance.


In addition to providing validity assurances, prediction regions attempt to be as efficient as possible; that is, be as small as possible while still valid. 
Density level sets, by construction, define the minimum sized region that contains 
a certain amount of probability mass \citep{Lei2013}. Additionally, modern advances in statistics have led to conditional density estimators (CDEs) $\widehat f(y|\x)$ that can 
handle complex data structures \citep[e.g.][]{Izbicki2017}. As such, it should be no surprise that CDEs have been increasingly used in conjunction with conformal inference \citep[e.g][]{Lei2014, Izbicki2021}.

\citet{Izbicki2021} recently proposed a 
method that aims to achieve local conformal coverage 
even when $\X$ is high-dimensional. To avoid the curse of dimensionality, 
they define local partitions of 
$\mathcal{X}$ 
based on the structure of the fitted CDEs $\widehat f$. 
This method obtains asymptotic conditional validity,
\[
\mathbb{P}_{(\X,Y) \sim F}(Y\in C(\X) |\X = \x) > 1 -\alpha\,,\ \forall \x \in \mathcal{X},
\]
but requires 
strong assumptions. 
These assumptions require relatively strong convergence of $\widehat{f}$ to the true conditional density for almost all $X$ values \citep[][Assumption 9]{Izbicki2021}.

This assumption of a well fit CDE $\widehat f$ may be unrealistic for finite-sample real world examples, 
especially in fields where models may be intentionally constrained by 
fundamental assumptions or selected from a less flexible
model class. 

In this work, we propose a 
novel approach that, while similar in spirit to 
\citet{Izbicki2021}, 
we believe gets closer to the heart of local conformal inference. Our method 
defines 
local partitions of $\mathcal{X}$ based on the local performance of a fitted CDE model. 
In particular, we define a distance between $\X$ values that reflects the difference between the true local coverage of CDE's level sets and expected coverage. This distance attempts to partition $\mathcal{X}$ into groups that would have similar conformal score distributions under an unattainable conditional conformal approach. 
Our approach is able to evaluate coverage by leveraging tools presented in \citet{zhao2021diagnostics}, a recent paper that explores the local validation of conditional density models. 

This paper is organized as follows. In Section \ref{sec:related_work}, we provide relevant background on conformal inference and model diagnostic tools. 
In Section \ref{sec:methods}, we introduce our proposed method for local conformal inference and describe its motivation. In Section \ref{sec:examples}, we present two examples to demonstrate that our method achieves local validity for well-designed partitions, even in high dimensions. Finally, in Section \ref{sec:discussion}, we discuss conclusions and future work. 

%% file: sections/related_work.tex


In this section, we 
provide relevant background on the foundations of conformal inference, approaches to local conformal inference, and local model diagnostics. 

The field of conformal inference 
aims to create prediction regions with valid coverage, under minimal parametric assumptions 
\citep{Vovk2005}. 
Conformal methods can be seen as a type of ``wrapper'' around black box models, generating valid, finite sample prediction regions while requiring only the assumption that previous and future observations are exchangeable. One creates conformal prediction regions by evaluating potential candidate values $y$, 
and including them if they are not 
too extreme relative to 
previous observations. The measure of extremeness comes from a conformal or non-conformal score. A conformal score has smaller values indicating more extremeness; a common example is the conditional density estimate $\widehat{f}(y|\x)$. A non-conformal score has larger values indicating more extremeness; a common example is the quantity $|y -\mu(\x)|$ defined using 
a regression function $\mu(\x)$.
(Multiplying by -1 would convert a conformal score to a non-conformal score, and vice-versa.) 

To illustrate 
how conformal inference works in more detail, we describe a popular method called split-conformal inference \citep[][pg. 110-111]{Vovk2005}. Assuming we have already define a conformal score function $cs(\cdot)$, split-conformal inference decides if a new $y$ should be contained in a prediction region by following three steps.
The first step is to split the observed data into disjoint training and calibration sets. We 
denote the indices for observations $i \in \{1,...,n\}$, 
and let the sets $\mathcal{I}_{train}$ and $\mathcal{I}_{cal}$ contain the indices of the training and calibration points, respectively. 
The second step is to build a prediction model using the training set, 
which is used to calculate conformal scores ($cs\left(y_i|\x_i, \{\x_j:j \in \mathcal{I}_{train}\}\right)$) for observations in the calibration set ($i \in \mathcal{I}_{cal}$).  
Finally, the third step is to 
evaluate potential $\widehat{y}_{n+1}$ values, and include these values in our prediction region if the conformal score 
$cs\left(\widehat{y}_{n+1}|\x_{n+1}, \{\x_j:j \in \mathcal{I}_{train}\}\right)$ is less extreme than $100\alpha\%$ of the conformal scores for the calibration set. Algorithm \ref{alg:split_conformal} formally presents 
the split-conformal approach. 

Split-conformal inference obtains finite sample validity 
as a result of the exchangeability of points in the calibration set and the new point. Given this exchangeability, the rank of the conformal score for the true $y_{n+1}$ with respect to the conformal scores 
from the calibration set 
is distributed uniformly between 1 and $1+|\mathcal{I}_{cal}|$. 
Therefore, including all $\widehat{y}_{n+1}$ that are less extreme than $100\alpha\%$ of the conformal scores for the calibration set creates a prediction region that a finite sample validity of $1-\alpha$ as desired.



\begin{algorithm}
	\caption{Split Conformal Inference}
	\label{alg:split_conformal}
	\begin{algorithmic}
		\State \textbf{Input}: {\small Conformal measure $cs$, significance level $1-\epsilon$, previous observed pairs $\{z_i\}_{i = 1}^n$ where $z_i = (\x_i, y_i)$, disjoint index sets for training and calibration sets ($\mathcal{I}_{train}$ and $\mathcal{I}_{cal}$), a new object $x_{n+1}$ and a potential $y$.}
		\State \textbf{Task}: 
		Decide if $y$ should be 
		included in a prediction region for $x_{n+1}$ at significance level $1-\epsilon$.
		\State \textbf{Algorithm}:
		\begin{enumerate}
			\item[1.] Provisionally view the $z_{n+1} = (x_{n+1}, y)$
			\item[2.1.] \textit{Conformal scores for the calibration data's set}:
			\item[] For all $j \in \mathcal{I}_{cal}$ set 
			\[
			\alpha_j = cs(z_j|\{z_i:i \in \mathcal{I}_{train}\}) 
			\]
			\item[2.2.] \textit{Conformal scores for new observation}: 
			\item[] \[
			\alpha_{n+1} = cs(z_{n+1}|\{z_i:i \in \mathcal{I}_{train}\}, ) 
			\]
			\item[3.] Set 
			\[
			p_{y} = \frac{\# \{j \in \mathcal{I}_{cal} : \alpha_j \geq \alpha_{n+1}\}}{|\mathcal{I}_{cal}| + 1}
			\]
			\item[4.] Include $y$ in the new prediction region if $p_y > \epsilon$.
		\end{enumerate}
	\end{algorithmic}
\end{algorithm}


As mentioned, 
classical conformal inference only provides marginal finite sample validity 
(that is, $\mathbb{P}_{(\X,Y) \sim F}(Y\in C(\x)) > 1 -\alpha$). 
However, 
practitioners are often more interested 
in 
conditional validity. It is known that distribution-free methods cannot achieve conditional coverage, unless one imposes strong assumptions on the underlying distribution \citep{Vovk2013, Lei2014, Barber2019a}. 
In this paper, we 
focus on local conformal inference. In local conformal inference as defined by \citet{Barber2019a}, validity is obtained in smaller partitions 
$A$ of the $\mathcal{X}$ space, such that $\mathbb{P}_{(\X,Y) \sim F}(Y\in C(\x) |\x \in A) > 1 -\alpha$ for each partition $A$. When used with split-conformal inference, this means that in the final step of creating a prediction region, we 
only view calibration points in the \textit{same} $A \subseteq \mathcal{X}$ as the new test $\X$ value as exchangeable with that new test observation, and therefore only use \textit{those} calibration points’ scores to define our 
notions of extremeness.
We now describe two different local conformal approaches, both of which 
mirror the structure that \citet{Barber2019a} describes and 
seek to define desirable local regions. \citet{Lei2014} defines 
local regions relative to the $X$ space, and creates bins of $X$ values based on Euclidean distance between points. 
As the number of observations 
increases, the size of the $X$ bins can decrease; in the limit of $n\to\infty$, 
asymptotic conditional coverage is achieved. 
\citet{Izbicki2021}’s \texttt{CD-split+} method provides a different way to create localized 
partitions. Concerned about the 
scalability of \citet{Lei2014}’s approach to high dimensional $\mathcal{X}$ spaces, 
\texttt{CD-split+} defines 
partitions of $\mathcal{X}$ by fitting a CDE model $\widehat f(y|\x)$ and 
using a similarity measure of the predicted conditional density estimates. Specifically, \texttt{CD-split+} creates local groups of $X$ values for those that are close relative to a profile distance:
\[
d_\text{profile}^{2}\left(\x_{a}, \x_{b}\right):=\int_{-\infty}^{\infty}\left(\widehat{H}\left(y \mid \x_{a}\right)-\widehat{H}\left(y \mid \x_{b}\right)\right)^{2} d y\;,
\]
where $\widehat{H}(y|\X)$ is the model estimated highest predictive density (HPD) value:

\begin{align}
    \label{eq:HPD}
    \HPD(y; \x) = \int_{{y'}:\widehat{f}({y'} | \x) \geq \widehat{f}(y | \x)} \widehat{f}({y'} | \x) d{y'}.
\end{align} 
This approach 
avoids the curse of dimensionality. Under the assumption that the estimated conditional density converges to the true conditional density, \texttt{CD-split+} 
obtains asymptotic conditional validity.
\citet{Izbicki2021} also proposes another technique called \texttt{HPD-split} (which is similar to \citet[][Appendix D]{Gupta2020}). 
\texttt{HPD-split} 
uses HPD values instead of CDE values as 
conformal scores, but still crucially relies on the assumption that $\widehat f$ is well fit. 


What happens to local conformal inference techniques when $\widehat f$ is not well fit? We explore this question in our paper, with the help of recently developed tools for diagnosing issues with CDE models. \citet{zhao2021diagnostics} 
has recently developed diagnostics for validating the goodness-of-fit of CDEs locally in the feature space $\mathcal{X}$. The crux of their method involves 
fitting a model to smoothly estimate the proportion of time an 
observation 
obtains an HPD 
value above a certain threshold, at any location $\X=\x$. In particular, they define a set of functions indexed by $\alpha \in [0,1]$:
\begin{equation}
    \label{eq:r_alpha}
    r_{\alpha}(\x) = \pr\left(\widehat{HPD}(Y;\x) < \alpha | \x\right)\;,
\end{equation}
where $\widehat{HPD}(\cdot)$ is the estimated HPD value function from the given CDE model $\widehat f(y|\x)$. Recall that HPD values would be uniform if the model were correctly fit and would also imply that $r_{\alpha}(\x) = \alpha$ for any $\alpha \in [0,1]$. 
\citet{zhao2021diagnostics} showed that we can estimate $r_{\alpha}$ by fitting a quantile regression model $\widehat r_{\alpha}$ across the feature space. By considering a grid of values for $\alpha \in [0,1]$, we obtain an informative sequence of $\{\widehat r_{\alpha}\}$ estimates, at any given location $\x \in \mathcal{X}$ in the feature space. Deviations of $\widehat r_{\alpha}(\x)$ from $\alpha$ for various values of $\alpha$ indicate that the CDE model $\widehat f$ does not fit the true conditional density $f$ well. Furthermore, the specific pattern of which values of $\alpha$ are in disagreement, and by how much, form a rough ``signature'' of the specific nature of the distributional deviation of $\widehat f$ from $f$.


%% file: sections/methods.tex

In this section, we propose a novel local conformal inference method, \texttt{MD-split+}, which synthesizes the above ideas in order to choose a better informed partition $\A$ of the feature space $\mathcal{X}$. Our method is practical to use and scales to high dimensions. 
In particular, it enables practitioners to leverage CDE models to create valid prediction regions, while being robust to misspecifications of those models.


\texttt{MD-split+} 
fits under the split-conformal framework.
It utilizes the following data splits:
(i) training set to build a CDE model $\widehat f$, (ii) a validation set to select optimal parameters for $\widehat f$, (iii) a diagnostic set (to be discussed in more detail) and (iv) the standard split-conformal calibration set.

Our procedure first trains a CDE model 
with the training and validation sets. Then, we train a quantile regression function $\{\widehat{r}_\alpha(\cdot): \alpha \in [0, \delta, 2\delta, \hdots, 1]\}$, to predict quantiles of HPD values 
for new points in the diagnostic set (as done in \citet{zhao2021diagnostics}). We also tune this quantile regression function using the validation set. (We imagine that the 
CDE model and quantile regression function are tuned in tandem, but one could also use 2 different validation sets.) 
Next, we define a \textit{model diagnostic 
distance}:
\[
d_\text{md}^{2}\left(\x_{a}, \x_{b}\right):=
\frac{1}{\lceil \frac{1}{\delta} \rceil + 1} \sum_{\substack{
\alpha \in \\ [0, \delta, 2\delta, \hdots, 1]}} \left(\widehat{r}_\alpha(\x_{a})-\widehat{r}_\alpha(\x_{b})\right)^{2}\;.
\]

Using this distance (in a similar fashion to \bpl{\citet{Izbicki2021}'s} \texttt{CD-split+}), 
we define $K$ 
clusters $C_1, \hdots, C_K$ to be the centroids of the calibration set points $X_i$ according to to $d_\text{md}^{2}$. 
These centroids define group indices for $\X$ values, via the index of the closest $C_k$ according to $d_\text{md}^{2}$. We can then apply local conformal inference, using this local structure combined with an HPD conformal score, to new test points. This procedure is 
detailed in Algorithm \ref{alg:md_split_plus}.

\begin{algorithm}
	\caption{MD-split+ Local Conformal Inference}
	\label{alg:md_split_plus}
	\begin{algorithmic}
		\State \textbf{Input}: {\small significance level $1-\epsilon$, previous observed pairs $\{z_i\}_{i = 1}^n$ where $z_i = (\x_i, y_i)$, disjoint index sets for training, validation, diagnostic and  calibration sets ($\mathcal{I}_{train}$, $\mathcal{I}_{val}, \mathcal{I}_{diag}, \mathcal{I}_{cal}$), and a new object $x_{n+1}$}.
		\State \textbf{Task}: create a prediction region for $y_{n+1}|\x_{n+1}$ at significance level $1-\epsilon$.
		\State \textbf{Algorithm}:
		\begin{enumerate}
			\item[1.] 
			Train conditional density estimator using $\{z_i: i \in \mathcal{I}_{train}\}$
			\item[2.] 
			Train model-fit quantile regression function $\{\widehat{r}_\alpha(\cdot)\}_\alpha$  using pairs $z_i' = \left(\widehat{HPD}(y_i|\x_i), \x_i\right)$ for $i \in \mathcal{I}_{diag}$.
			\item[3.] Tune conditional density estimator $\widehat{f}(\cdot)$ and quantile regression function $\{\widehat{r}_\alpha(\cdot)\}_\alpha$ parameters with $z_j$ and $z_j'$ for $j \in \mathcal{I}_{val}$ respectively.
			\item[4.] 
			Define grouping centroids by using \texttt{k-means++} on a discretized set of vector $[r_\alpha(\x_l) \text{ for } \alpha \in [0,...,1]]$ using $l \in \mathcal{I}_{cal}$.
			\item[5.] 
			Apply local split-conformal inference using the clustering defined in Step 4, 
			with conformal scores defined by $\widehat{HPD}(\cdot)$ with significance level $1-\epsilon$.
		\end{enumerate}
	\end{algorithmic}
\end{algorithm}

\subsection{Motivation and Intuition}

We believe that our approach, while similar in spirit to \citet{Lei2014} and \citet{Izbicki2021}, gets closer to the heart of local conformal inference.

A conformal prediction region can be viewed as a level set of the conformal score function on $\mathcal{Y}$. Specifically, the region contains all $y_{new}'$ such that $cs(y_{new}'|\x_{new})$ is greater than or equal to some threshold. In this way, a given conformal score function can be seen to define a collection of nested level sets \citep[][pg. 9]{Vovk2005}. 
Thus, if a confidence level of $1-\alpha$ is desired, conformal inference can be viewed as selecting the level set among this collection that would be expected to contain a probability mass 
of $1-\alpha$ (at least in a marginal sense).

Taking this view of conformal inference, if the conformal score were an HPD function, then 
conformal prediction regions would be 
level sets indexed by HPD values (as thresholds). If the HPD function (derived from the CDE model $\widehat f$) were well fit and we wanted a prediction region with confidence level $1-\alpha$, then conformal inference would 
simply return the set defined by $y$ values with HPD values $\geq 1-\alpha$. That is, we would obtain $\{y: HPD(y|\x) \geq 1-\alpha\}$. 
If the HPD function were not well fit, 
then conformal inference might select an HPD threshold different from the \textit{nominal} $1-\alpha$ level, in order to return a prediction set with \textit{achieved} $1-\alpha$ coverage. In 
this sense, conformal inference serves as 
a correction to the HPD function, which may be imperfectly estimated.

This motivates 
how our proposed \texttt{MD-split+} method partitions $\mathcal{X}$. 
\texttt{MD-split+} defines local partitions of $\mathcal{X}$ based on the local performance of a fitted CDE model. 
Effectively, what we aim to do is to group together $\X = \x$ values that would get similar ``corrections'' to their HPD function, if we were able to run 
a conditional conformal 
method at every $\X = \x$. In reality, of course, 
we cannot create a conditional conformal method. 
But we 
can estimate what those ``corrections'' would look like, at any location $\x \in \mathcal{X}$, thanks to the diagnostic framework of \citet{zhao2021diagnostics}. 
We 
use these estimates to form better local partitions of $\mathcal{X}$, 
pairing together $\X$ values that would see similar conformal score distributions under a conditional conformal method.

%% file: sections/examples.tex
\subsection{Simple Linear Regression Conditional Density Estimation}\label{sec:example1}



\begin{figure}
    \centering
    \includegraphics[width = 1\linewidth]{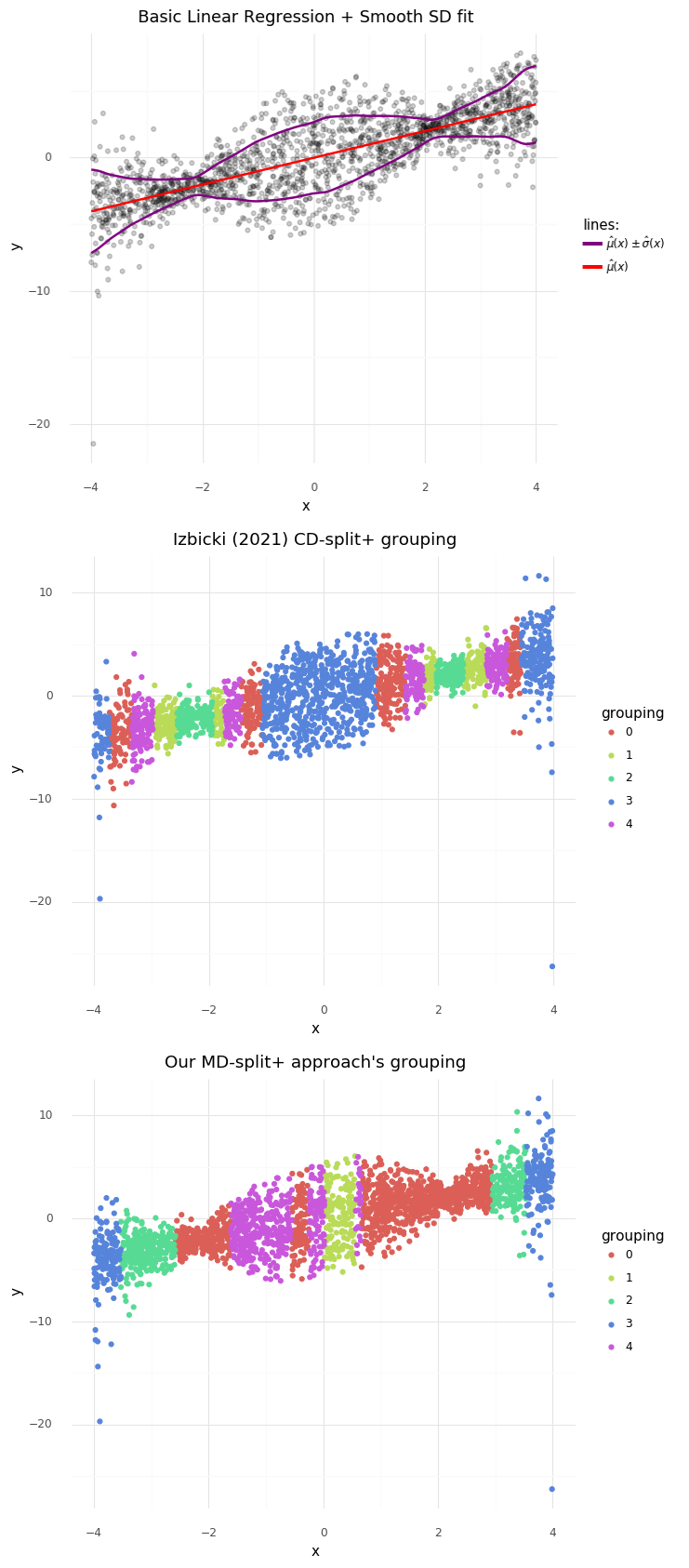}
    \caption{This figure relates to the example discussed in Section \ref{sec:example1}. The top subplot presents the underlying data distribution, as well as the fit of the mean and standard deviation that defines the CDE model. The lower two subplots display the local groupings for the local conformal approaches \texttt{CD-split+} and \texttt{MD-split+} 
    with 5 clusters each. Notice that 
    \texttt{CD-split+} groups $X$ values with $Y$ that have a truncated normal together with $X$ values with $Y$ that have $t$-distributions with low degrees of freedom, and in general groups by $\sigma(x)$ values ($|X|-2$). This is not optimal, given that they have largely differing conformal score distributions. \texttt{MD-split+} focuses more on grouping similar underlying distributions together. 
    }
    \label{fig:iz_v_us_linear_1d_left}
\end{figure}

Our first example 
has one-dimensional $\mathcal{X}$ and $\mathcal{Y}$ spaces, 
and we specifically developed it to be juxtaposed against 
\citet[][Figure 3]{Izbicki2021}. 
We define the underlying distribution of the data by $X \sim \text{Unif}(-4,4)$ and a varying $Y|X=x$ distribution 
centered at $x$. 
The conditional density distribution varies from a scaled $t$-distribution ($df=3$), to a scaled normal distribution, to a scaled truncated normal distribution, and back as $x$ varies between $-4$ and $4$. More specifically, the underlying distribution is defined as: 

\vspace{2cm}
\begin{align*}
    X & \sim \text{Unif}(-4,4)
\end{align*}
If $X=x$ and $|x| \leq 2$:
\begin{fleqn}
\begin{align*}
		\quad (Y-x)  & \sim
			\begin{array}{l}
				\text{trunc\_norm}\Big(\sigma\!: \sigma(|x|-2),\\
				\quad  \text{low}\!:-\sigma(|x|-2)\cdot\text{trunc\_max}(|x|-2),\\
				\quad \text{up}\!: \sigma(|x|-2)\cdot\text{trunc\_max}(|x|-2)\Big) \end{array}
\end{align*}
\end{fleqn}
If $X=x$ and $|x| > 2$:
\begin{fleqn}
\begin{align*}
		\quad (Y-x) & \sim \frac{1}{2} \sigma(|x|-2) \cdot \text{student\_t}(\text{df}: \text{df}(|x|-2))
\end{align*}
\end{fleqn}

where

\begin{align*}
    \sigma(x) & :=  1 + 1.5 |x| \\
	\text{trunc\_max}(x) & := .5 + \log(2/|x|) \\
	\text{df}(x) & := (3-|x|)^3 + 2\,.
\end{align*}
Figure \ref{fig:iz_v_us_linear_1d_left}'s top subplot presents a set of random draws from this distribution.


We then fit a Gaussian with 
varying mean and 
spread to estimate the conditional density function 
for this $Y|X$. This is a modeling choice that a practitioner might reasonably try when presented with this data. However, this CDE model is inherently misspecified.

We 
compare the performance of our \texttt{MD-split+} 
with \citet{Izbicki2021}'s \texttt{CD-split+} on this example, and 
also discuss how \citet{Lei2014}'s local approach would have performed.

First, we note that \texttt{MD-split+} uses data
differently than 
\texttt{CD-split+} and \citet{Lei2014}'s local conformal approach would. 
In this example, \texttt{MD-split+} places
25\% in a training set, 25\% in validation set, 25\% in a diagnostic set and 25\% in calibration set, while 
\texttt{CD-split+} splits the data 33\% in training,
33\% in validation, and 33\% in the calibration set.

For both procedures, we separately fit 
a Gaussian conditional density estimator, where the conditional mean $\widehat{\mu}(\x)$ comes from a linear regression, and the conditional variance $\widehat{\sigma}^2(\x)$ comes from a smoothing spline fit to the squared residuals of the linear regression. 
Both fits (whether using 25\% data splits in \texttt{MD-split+} or 33\% data splits in \texttt{CD-split+}) are basically the same, and we visualize these two functions from \texttt{MD-split+} 
in Figure \ref{fig:iz_v_us_linear_1d_left}, top panel. 
For \texttt{MD-split+}, we also fit a smooth quantile regression function on HPD values computed using 
the CDE model, with a small
quantile regression neural network with Adam optimization \citep{kingma2014adam} with 
learning rate $10^{-3}$, and 2 hidden layers with 10 nodes each. One can imagine that a set of smoothing spline binary regressors would perform similarly across this simple space.

For both procedures we create 5 partitions of the $\mathcal{X}$ space, and these can be seen in Figure \ref{fig:iz_v_us_linear_1d_left}'s lower two subplots 
This figure shows that the \texttt{CD-split+} clusters utilizes the estimated $\hat{\sigma}(x)$ values (visually clustering by $|\x|$), whereas our approach uses the relative fit of the CDE model (visually clustering by $|\x|-2$). If we had also applied \citet{Lei2014}'s approach, we would have seen clusters based solely on $\X$ value. Compared to \texttt{CD-split+}, our \texttt{MD-split+} algorithm clusters with respect to the underlying distributional differences the data has. Compared to \citet{Lei2014}'s, \texttt{MD-split+} clusters more similar underlying residual distributions, by not being constrained to only cluster linearly on contiguous $\X$ values. This simple example highlights a drawback of
clustering on model conditional density estimation when model fit is imperfect. 

\begin{figure}
    \centering
    \includegraphics[width = 1\linewidth]{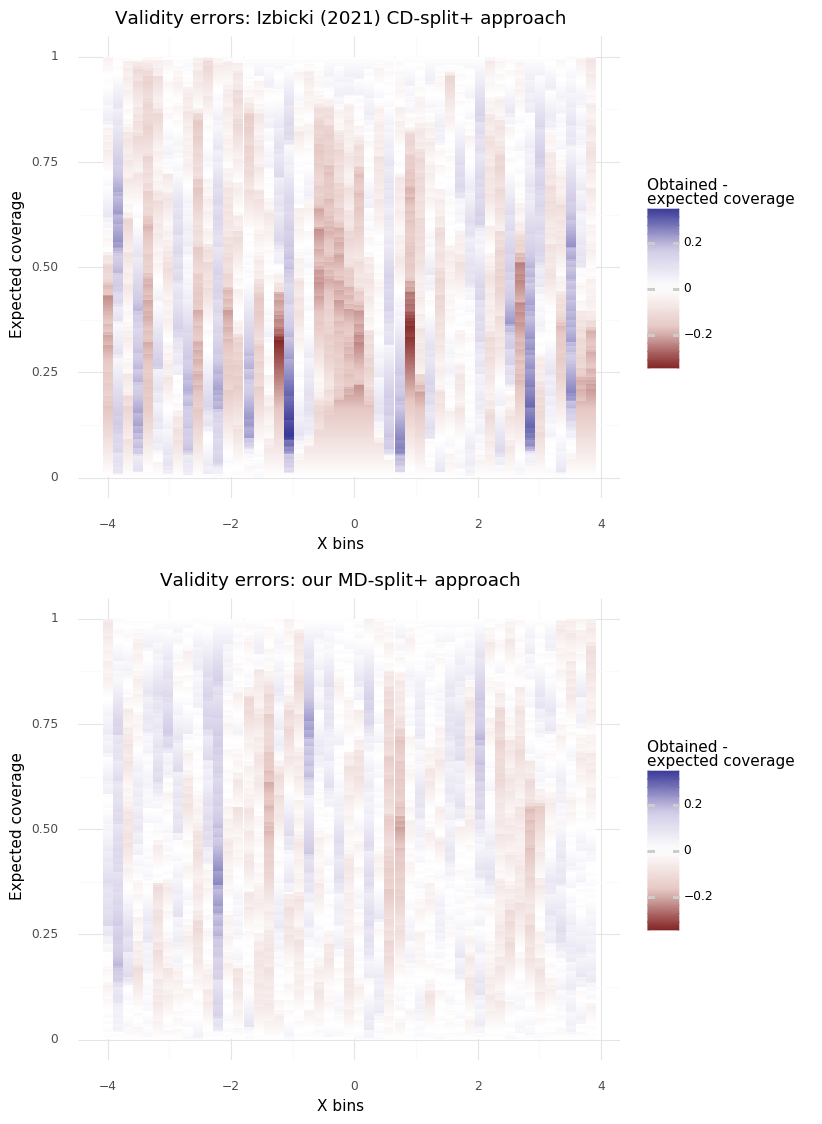}
    \caption{This figure captures the difference in empirical coverage and expected coverage  for small bins on the $X$ space related to the generative distribution for example 1 and the clusters visualized in Figure \ref{fig:iz_v_us_linear_1d_left}. The top subplot relates to the \texttt{CD-split+} approach with 5 clusters and the lower subplot relates to our proposed approach with 5 clusters as well. 
    }
    \label{fig:iz_v_us_linear_1d_right}
\end{figure}

In addition, this misspecified CDE model (Gaussian with varying mean and variance) means that as the number of observations ($n$) and the number of clusters ($k$) increases, we will \textbf{not} see asymptotic conditional validity emerge under \texttt{CD-split+}. 
Figure \ref{fig:iz_v_us_linear_1d_right} captures the difference in local validity for a grid of small bins of $X$ values relative to both clustering approaches, with stronger colors highlighting large differences. As expected, \texttt{CD-split+} performs worse on this localized level than our \texttt{MD-split+} approach. In other words, a prediction region made using \texttt{CD-split+} will more often have a different amount of conditional coverage than desired, and this difference will often be larger than that seen with prediction regions made using \texttt{MD-split+}.

\subsection{Convolutional Neural Density Estimation for Galaxy Images}\label{sec:example2}

\begin{figure}
    \centering
    \includegraphics[width = 1.1\linewidth]{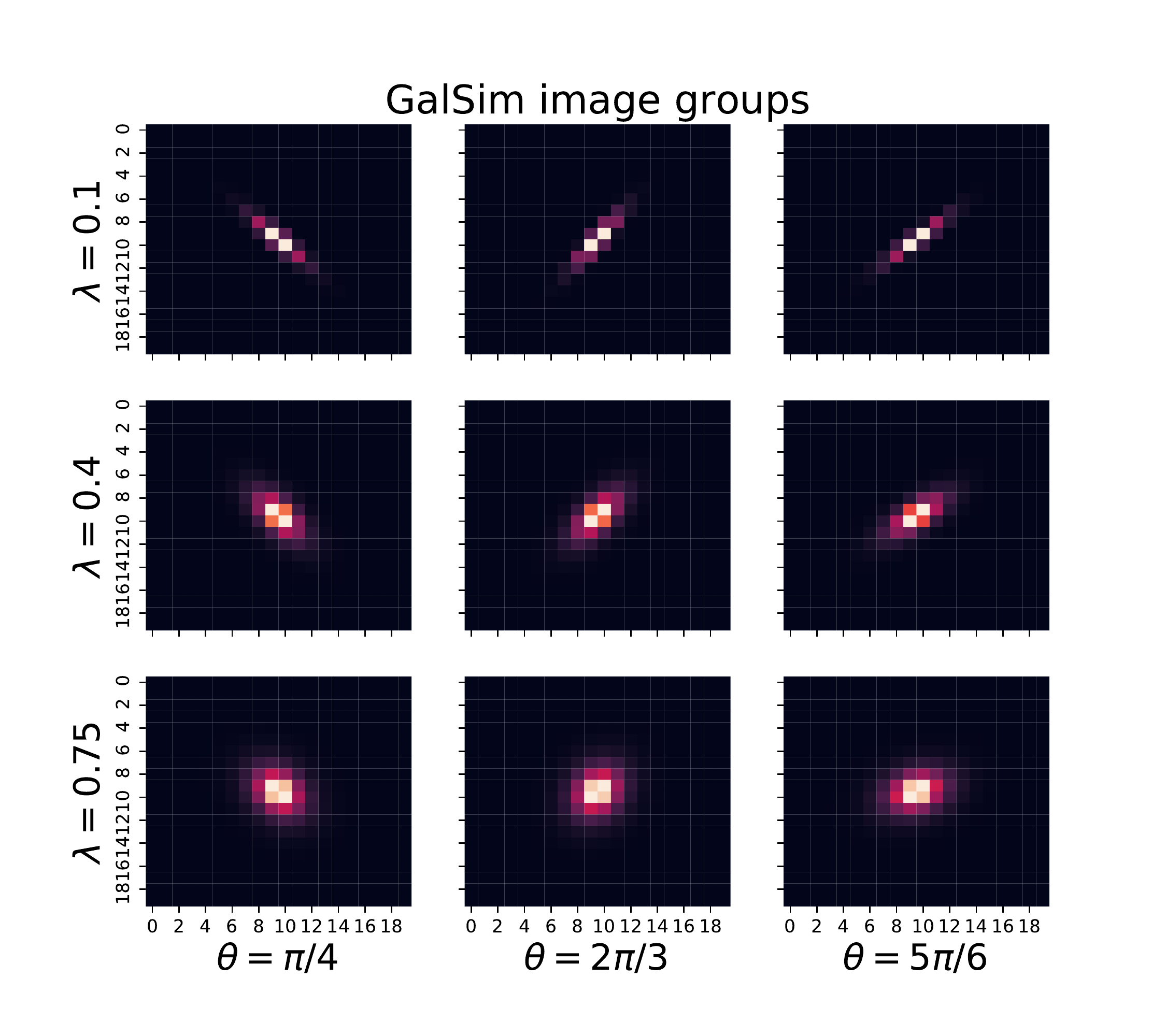}
    \caption{We create 9 equally sized populations of galaxy images, representing combinations of axis ratios $\lambda\in \{0.1,0.4,0.75\}$ and angles $\theta\in \{\pi/4, 2\pi/3, 5\pi/6\}$. Individual images have a small amount of noise in both $\lambda$ and $\theta$. This grid shows a representative image from each group.}
    \label{fig:gal_imgs}
\end{figure}

In this example, we consider the problem of estimating 
prediction regions for synthetic ``redshift'' values $Z$ 
(the response), which are assigned to photometric or “photo-z” galaxy images $\X$ (the features). 
Here, 
$\X$ represents a $20 \times 20$-pixel image of an elliptical galaxy generated by \texttt{GalSim}, an open-source toolkit for simulating realistic images of astronomical objects \citep{rowe2015galsim}. In \texttt{GalSim}, we can vary the axis ratio $\lambda$, defined as the ratio between the minor and major axes of the projection of the elliptical galaxy, as well as the galaxy's rotational angle $\theta$ with respect to the x-axis. We create 9 equally sized populations of galaxies, representing every combination of $\lambda \in \{0.1, 0.4, 0.75\}$ and $\theta \in \{\pi/4, 2\pi/3, 5\pi/6\}$\footnote{Note that these $\theta$ values are not equally spaced.}, with uniform noise added to $\lambda$ and $\theta$ for each individual image. See Figure \ref{fig:gal_imgs} for representative examples of each galaxy image type.

\begin{figure}
    \centering
    \includegraphics[width = 1.1\linewidth]{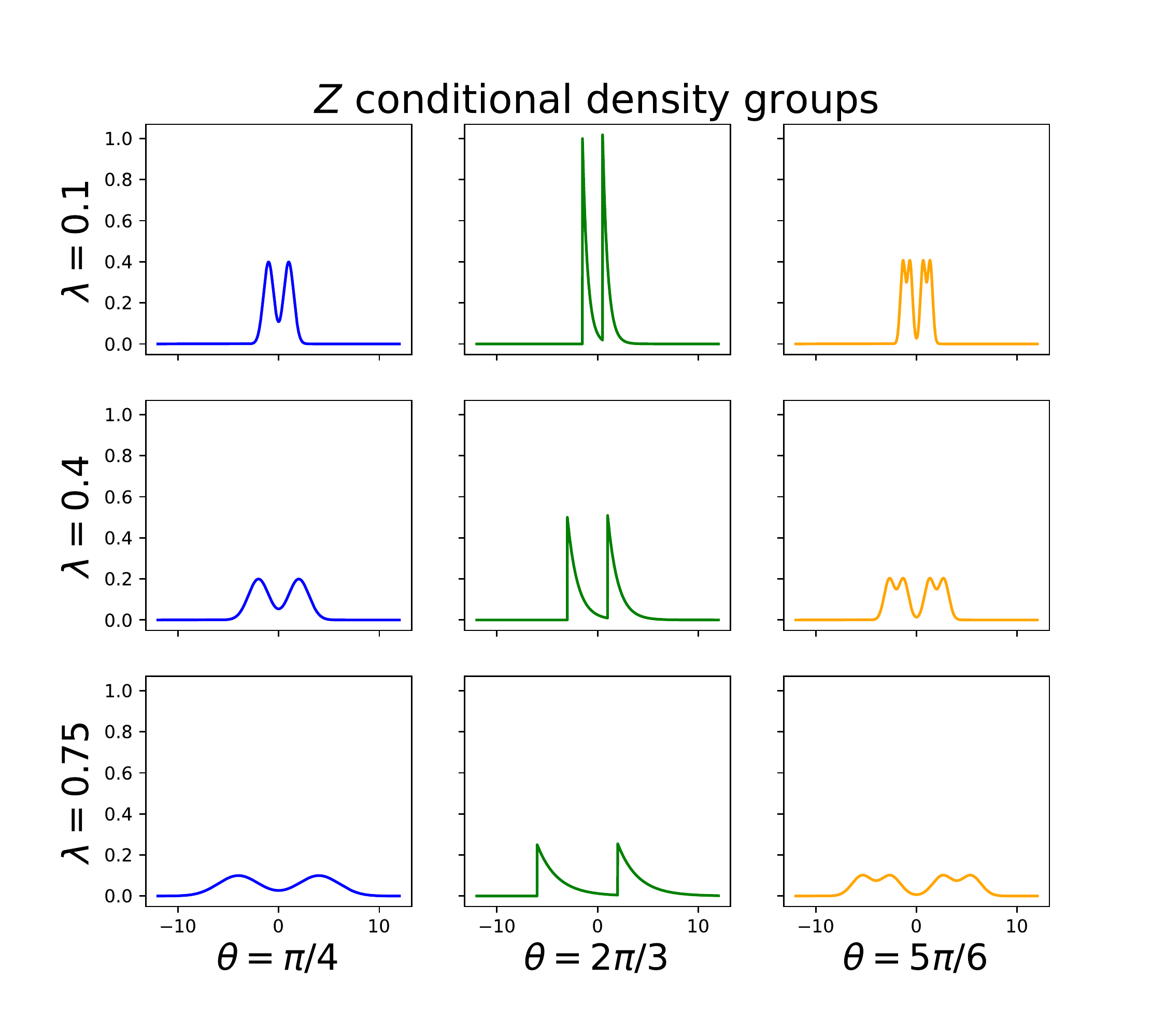}
    \caption{We define 9 conditional density models for the ``redshift'' $Z$, corresponding to the 9 galaxy image populations in Figure \ref{fig:gal_imgs}. Galaxy images with the same $\lambda$ value have densities $f(z|\x)$ that have the same variance, while those with the same $\theta$ value have densities $f(z|\x)$ that have the same distributional shape.}
    \label{fig:z_dists}
\end{figure}

We then assign a response variable $Z$ 
with 
different bimodal mixture distributions as follows:
\begin{align*}
    Z|\lambda,\theta \sim a_{\lambda} \left[ \frac{1}{2} U_{\theta} + \frac{1}{2} V_{\theta} \right]
\end{align*}
where
\begin{align*}
    U_{\theta} = 
    \begin{cases}
        \mathcal{N}(2,1), & \theta=\pi/4 \\
        \text{Exponential}(1) \text{ centered at 2}, & \theta=2\pi/3 \\
        \frac{1}{2}\mathcal{N}(2.7,0.25) + \frac{1}{2}\mathcal{N}(1.3,0.25), & \theta=5\pi/6
    \end{cases}
\end{align*}
\begin{align*}
    V_{\theta} = 
    \begin{cases}
        \mathcal{N}(-2,1), & \theta=\pi/4 \\
        \text{Exponential}(1) \text{ centered at -2}, & \theta=2\pi/3 \\
        \frac{1}{2}\mathcal{N}(-2.7,0.25) + \frac{1}{2}\mathcal{N}(-1.3,0.25), & \theta=5\pi/6
    \end{cases}
\end{align*}
and 
\begin{align*}
    a_{\lambda} = 
    \begin{cases}
        0.5, & \lambda=0.1 \\
        1, & \lambda=0.4 \\
        2, & \lambda=0.75
    \end{cases}
\end{align*}
See Figure \ref{fig:z_dists} for a plot comparing these distributions.

Based on the marginal distribution of the response, 
a practitioner might reasonably decide to 
model conditional density estimates as a mixture of Gaussians. 
In particular, here we fit a Gaussian convolutional mixture density network (ConvMDN, \cite{disanto2018cmdn}), which can effectively train on 
the image feature space.

This is a situation where the data structure is 
more complex, so local conformal inference is challenging. 
\citet{Lei2014}'s partitioning method does not scale well here because $\X$ is high-dimensional. Furthermore, we will demonstrate that even well-tuned, sophisticated conditional density models trained on this dataset with reasonably large sample sizes may be misspecified in subtle but fundamental ways. Thus, the methods of \citet{Izbicki2021} may not lead to sensible local subgroups.
On the other hand, our proposed \texttt{MD-split+} method provides a practical way to do local conformal inference successfully, even on complex data with 
imperfectly fit CDEs. 
In this example, we 
compare \texttt{MD-split+} to the related approach \texttt{CD-split+} \citep{Izbicki2021}. 

For this complex data structure, we dedicate more data to training and validating the ConvMDN model. Specifically, for our \texttt{MD-split+} implementation, we dedicate 27,000 observations to the training and validation sets (70/30 split), 4,500 observations for our diagnostic set (to train our HPD quantile functions), and 4,500 for the calibration set. For \texttt{CD-split+} we add what was the diagnostic set above to the training and validation sets, thus giving us 31,500 observations (again with 70/30 split). We also evaluate both approaches on 4,500 test observations.

For both approaches, 
we fit a Gaussian convolutional mixture density network (ConvMDN, \cite{disanto2018cmdn}), with two convolutional and two fully connected layers with ReLU activations \citep{glorot2011relu}. We train 
these models using the Adam optimizer \citep{kingma2014adam} with learning rate $10^{-3}$, $\beta_1 = 0.9$, and $\beta_2 = 0.999$. 
We tune several hyperparameters in search of the best fitting ConvMDN model. We allow $M$, the number of mixture components, to vary from 2 to 10. We also tune the number of hidden units in the penultimate layer and whether to include dropout \citep{srivastava2014dropout} after the convolutional layers. 
For both approaches, we found that the best ConvMDN model with the lowest KL loss had parameters $M=2$, 10 hidden layers, and a 50\% dropout layer. 
For \texttt{MD-split+} we also train a quantile regression model (with $\alpha \in \{0, 0.02, ...., 1\}$) estimating $\widehat{HPD}(Y|\X)$ values given $\X$, where $\widehat{HPD}$ is defined relative to the ConvMDN.

\begin{figure}
    \centering
    \includegraphics[width = 0.8\linewidth]{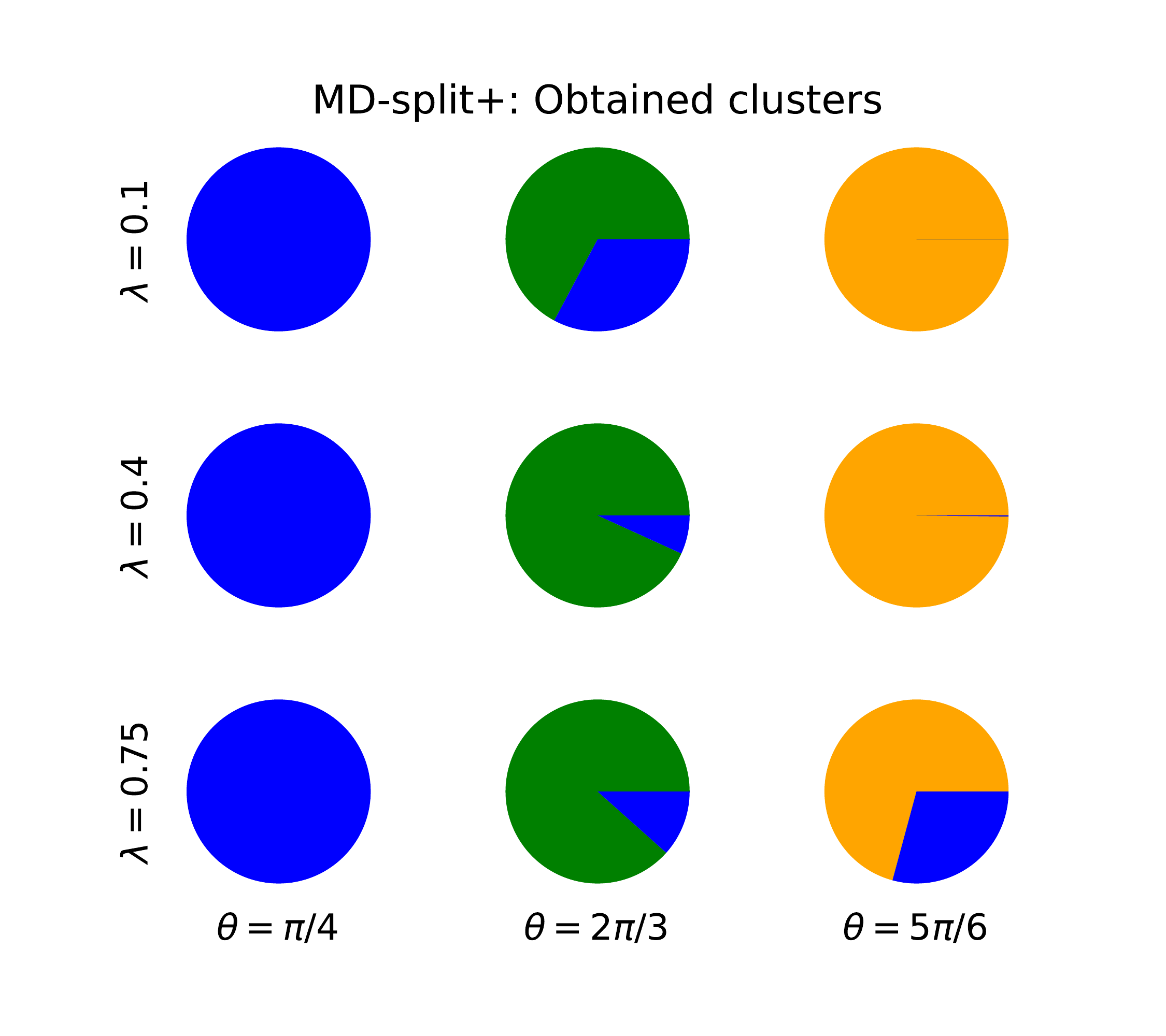}
    \includegraphics[width = 0.8\linewidth]{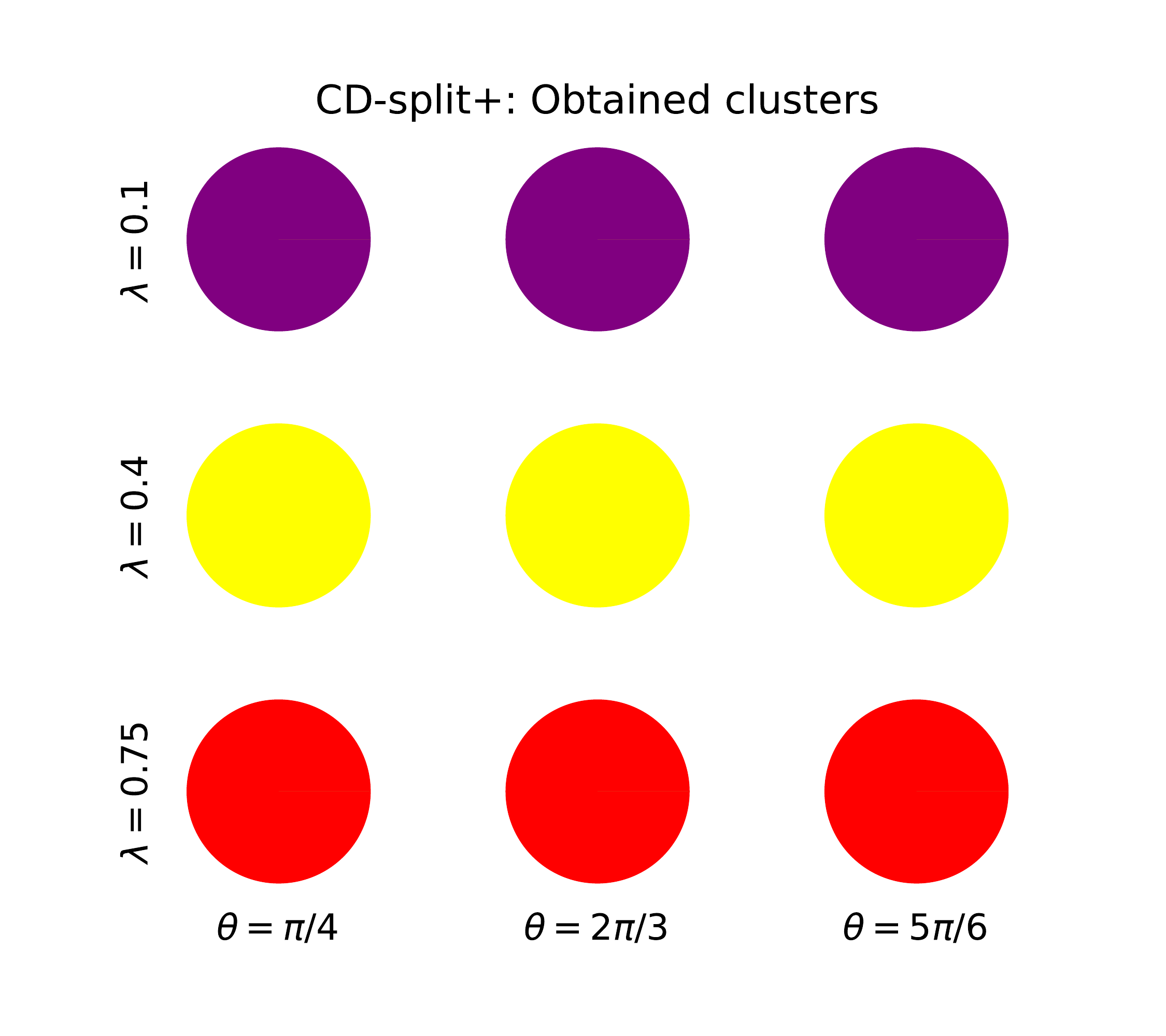}
    \caption{Pie charts showing the fraction of points assigned to different clusters. \textit{Top:} Our method \texttt{MD-split+} obtains reasonable clusters. For the most part, points with similarly shaped conditional densities (see Figure \ref{fig:z_dists}) are grouped together. Intuitively, such points will have similar HPD coverage profiles, and thus should belong to the same local conformal partition. \textit{Bottom:} \texttt{CD-split+} forms unintuitive clusters. This method clusters together distributions with the same variance, regardless of how differently shaped those distributions might be. 
    }
    \label{fig:clusters_grid}
\end{figure}

We then define three clusters of the calibration points using $d_{md}$ and $d_{profile}$ for \texttt{MD-split+} and \texttt{CD-split+} respectively. Next, we construct prediction regions on a test set of 4500 new observations (galaxy images + redshifts), and verify their coverage (whether the prediction region includes the true redshift). By construction, even this ``best'' ConvMDN model is misspecified, because the true conditional densities $f(z|\x)$ are not all mixtures of 2 Gaussians. Therefore, simply using 
a distance that compares fitted CDE models $\widehat f$,  
as \texttt{CD-split+} does, can lead to undesirable clusters that fail to achieve local conformal coverage on the true subpopulations of interest. In contrast, our method partitions the feature space into sensible regions, as 
the HPD distributions are the same across the scalings defined by $\lambda$. 
See Figure \ref{fig:clusters_grid} for a comparison of the clusters achieved by these two methods.

\begin{figure}
    \centering
    \includegraphics[width = 1.1\linewidth]{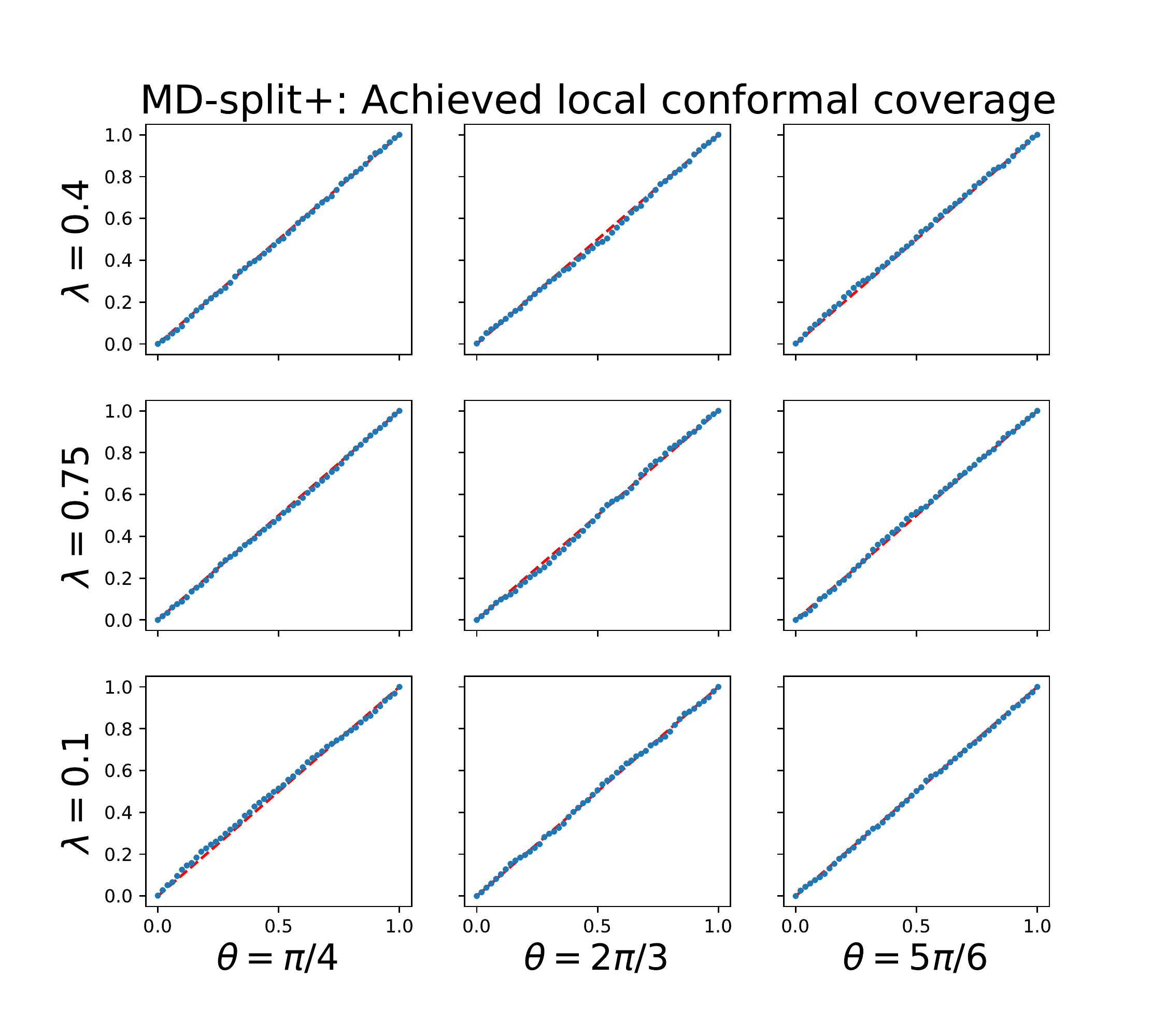}
    \includegraphics[width = 1.1\linewidth]{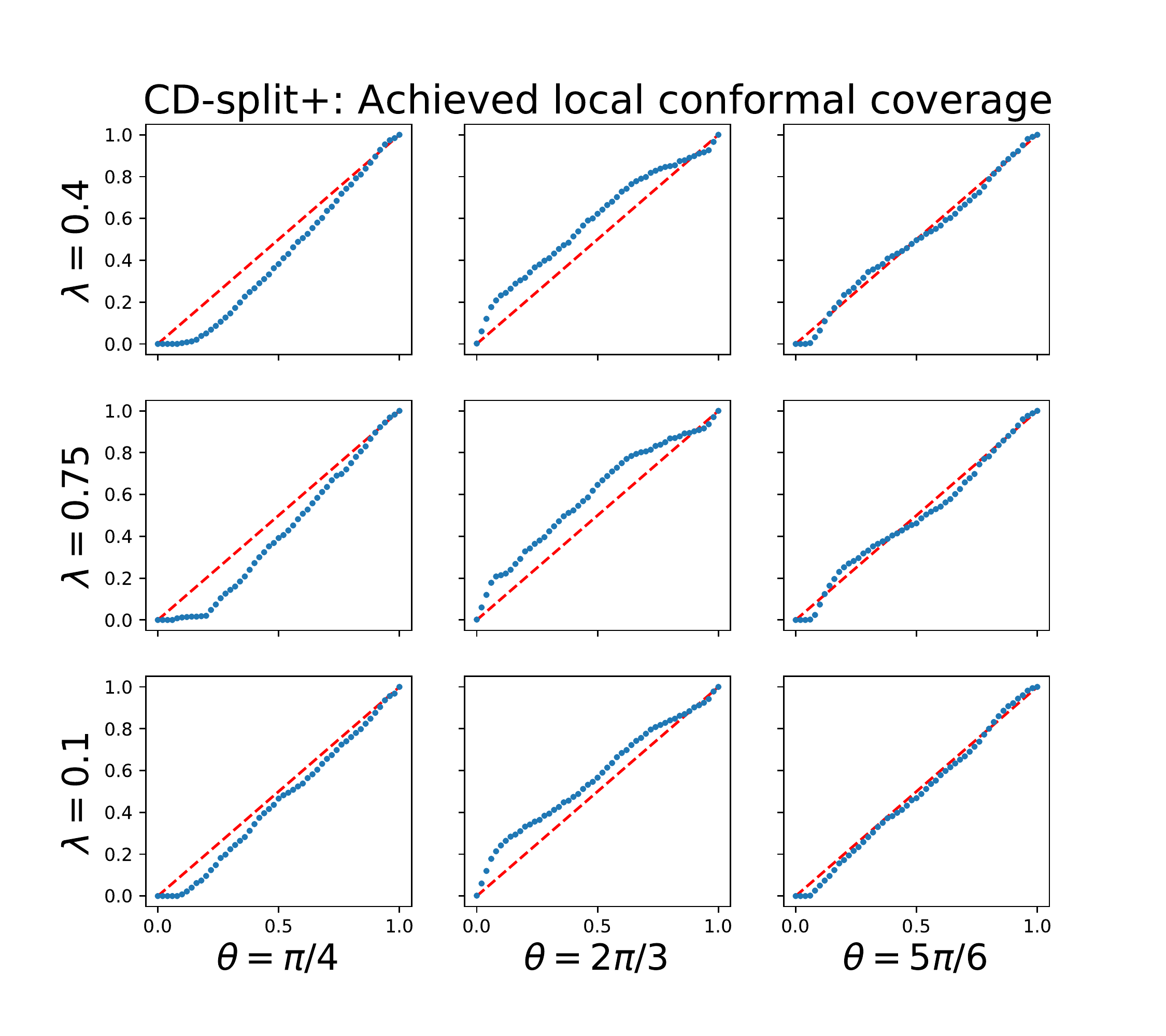}
    \caption{
    We check nominal vs. achieved coverage of local conformal sets within each of the 9 true subpopulations defined earlier. \textit{Top:} Using the 3 clusters obtained from our method \texttt{MD-split+}, we achieve correct local conformal coverage across the range of $\alpha\in[0,1]$. \textit{Bottom:} Using the 3 clusters obtained from \texttt{CD-split+}, we see clear deviations from correct local conformal coverage in many of the 9 true subpopulations.}
    \label{fig:conformal_QQ_plots}
\end{figure}

\begin{figure}
    \centering
    \includegraphics[width = 1.0\linewidth]{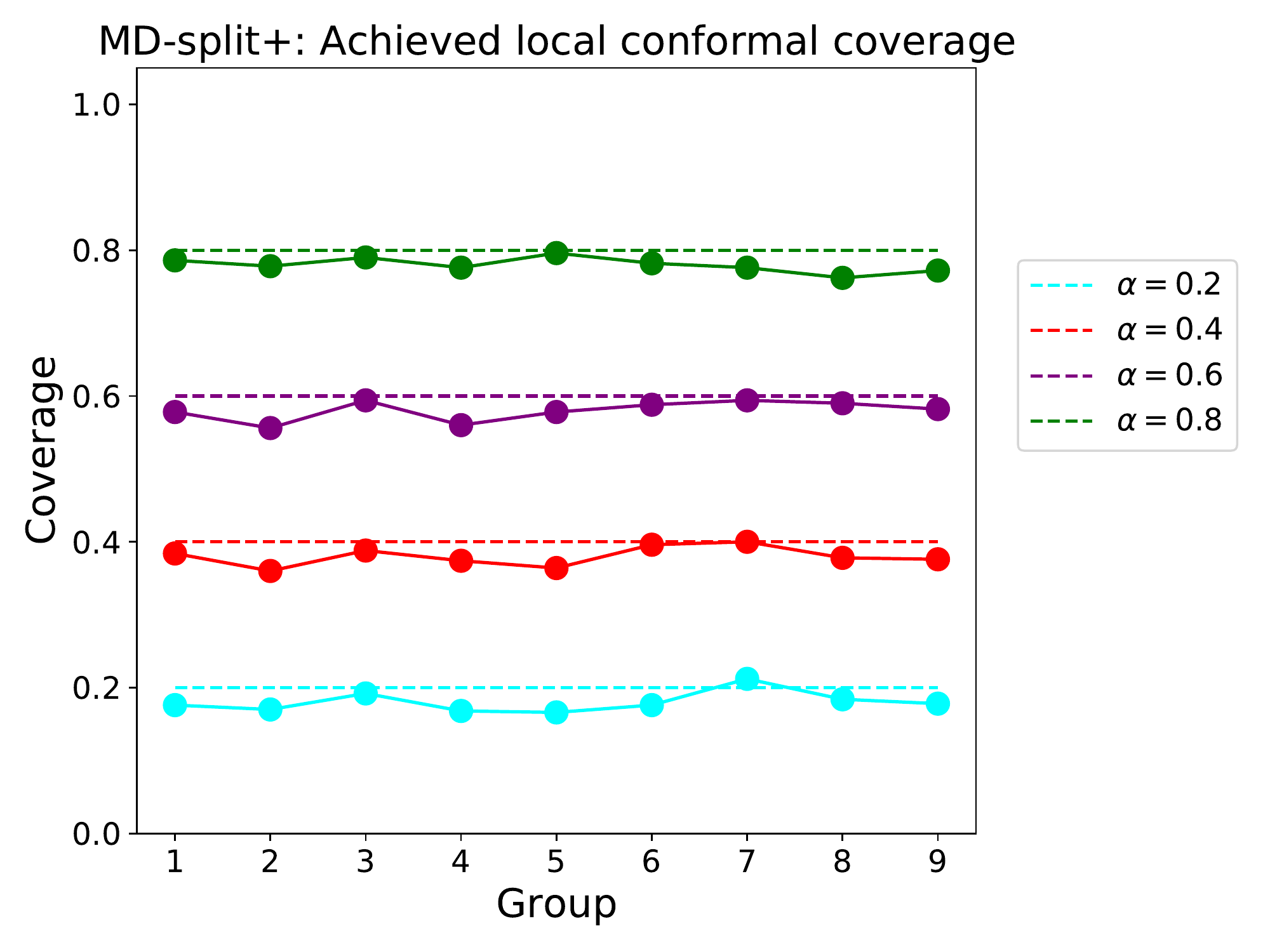}
    \includegraphics[width = 1.0\linewidth]{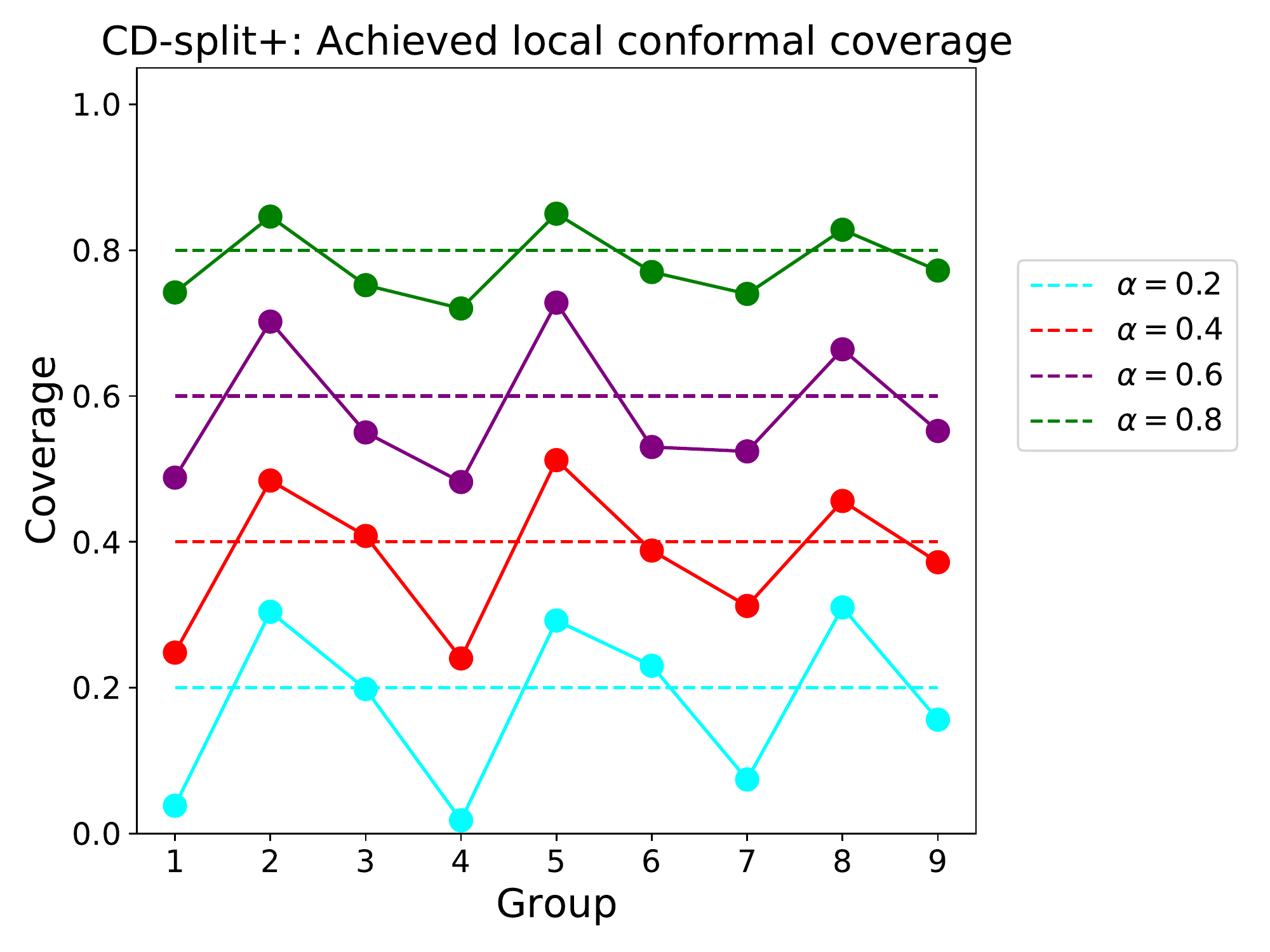}
    \caption{In this figure we select 4 particular confidence levels and check the achieved local conformal coverage by group, for nominal coverage levels of $\alpha=0.2,0.4,0.6,0.8$. \textit{Top:} Our method \texttt{MD-split+} achieves approximately correct coverage in every true subpopulation. \textit{Bottom:} \texttt{CD-split+} leads to large deviations from the correct coverage in most subpopulations. \texttt{CD-split+} clusters together groups (1,2,3), (4,5,6), and (7,8,9) in the figure. While coverage is correct when averaged across these clusters, it is incorrect for the true groups of interest. 
    }
    \label{fig:conformal_coverage_contours}
\end{figure}

Next, we form local conformal prediction regions using the two sets of clusters and their associated conformal scores. Recall that \texttt{CD-split+} uses conditional density estimate values, while \texttt{MD-split+} uses $\widehat{HDP}$ values. 
Of course, by construction one will obtain validity on average over each cluster, for any arbitrary clusters one wishes to define. But we are interested in whether 
validity is achieved over the 9 \textit{true} subpopulations in the dataset. Figure \ref{fig:conformal_QQ_plots} shows the empirical coverage relative to the conformal confidence levels 
across the 9 true subpopulations. 
We see that, across the range of confidence levels, our method obtains empirical coverage very close to what is expected. On the other hand, \texttt{CD-split+}'s empirical coverage for these 9 true subpopulations varies drastically. 
Figure \ref{fig:conformal_coverage_contours} 
shows this information for four specific confidence levels (0.2,0.4,0.6, and 0.8), again highlighting the performance differences between these two methods. 

In this example, we have seen how a complex CDE model $\widehat f$ can remain misspecified, despite careful tuning. Moreover, this kind of misspecification would not be obvious to the practitioner. In theory, one can always simply ``find a better model'', but in reality, models are often imperfect, especially in complex data settings. It is in this realistic context that \texttt{MD-split+}, unlike previous known methods, can still achieve valid, scalable local conformal inference.

%% file: sections/discussion.tex
We have developed a novel local conformal inference method that is both scalable to high dimensions and extendable 
to complex data settings where conditional density estimators may be poorly estimated. 
As such, our method has great practical utility. If CDE models were always well fit, then the estimated conditional density level sets would always be close to the true level sets, and there would be no reason to use conformal inference at all.
Instead, one could simply use the CDE model to select the level set with the probability mass equal to the desired confidence level.

In comparison with other local conformal inference techniques that use CDE or HPD values as their conformal scores, our approach does require an additional data split. However, the example in Section \ref{sec:example2} shows that this added split will often not be too burdensome for the practitioner. In developing our approach, we observed the impact of having the CDE model or HPD quantile regression model not being well tuned. A poorly tuned CDE model (specifically one that is overfit), can make it harder for the HPD quantile regression function to correctly cluster $\X$ observations based on model performance. Similarly, a poorly tuned HPD quantile regression function can reduce the ability to obtain optimal clusters.

There is a lot of potential for future work in understanding the theoretical aspects and more nuanced properties of our approach. Specifically, we are interested in exploring whether our method achieves asymptotic conditional coverage with meaningfully weaker assumptions than, say, \citet{Izbicki2021}. We expect this to require understanding the smoothness in $\mathcal{X}$ of the true and estimated conditional density, as well as the smoothness in $\mathcal{X}$ of the oracle versus estimated HPD quantile regression function. 
Additionally, we seek to better understand (potentially empirically) how comparisons of our \texttt{MD-split+} (with less data to train the estimated CDE model) to \citet{Izbicki2021}'s \texttt{CD-split+} might change if, instead of fundamentally misspecified models, we had CDE and HPD quantile regression functions that were poorly fit due to finite sample sizes, but eventually converge to the true conditional density and oracle HPD quantile regression functions, respectively.

Overall, we see \texttt{MD-split+} as a new tool in the practitioner's toolkit for performing local conformal inference using conditional density estimation$-$one that is especially useful when $\mathcal{X}$ is high dimensional or the CDE is misspecified. We hope that our procedure will appeal to a wide range of practitioners, and that they will appreciate the philosophical foundations of our approach as one of its strengths. 

%% file: sections/acknowledgements.tex

BL and DZ are grateful for fruitful 
discussions with Aaditya Ramdas about this topic. 
BL would like to thank his advisor 
Chad Schafer 
for discussions on local conformal inference literature and potential extensions. DZ would like to thank his advisor 
Ann B. Lee 
and 
collaborator 
Rafael Izbicki, with whom he developed the conditional density model diagnostic tools leveraged in this paper.

%% file: main.bbl
\begin{thebibliography}{15}
\providecommand{\natexlab}[1]{#1}
\providecommand{\url}[1]{\texttt{#1}}
\expandafter\ifx\csname urlstyle\endcsname\relax
  \providecommand{\doi}[1]{doi: #1}\else
  \providecommand{\doi}{doi: \begingroup \urlstyle{rm}\Url}\fi

\bibitem[Barber et~al.(2019)Barber, Cand{\`{e}}s, Ramdas, and
  Tibshirani]{Barber2019a}
Rina~Foygel Barber, Emmanuel~J. Cand{\`{e}}s, Aaditya Ramdas, and Ryan~J.
  Tibshirani.
\newblock {The limits of distribution-free conditional predictive inference}.
\newblock mar 2019.
\newblock URL \url{http://arxiv.org/abs/1903.04684}.

\bibitem[D’Isanto and Polsterer(2018)]{disanto2018cmdn}
Antonio D’Isanto and Kai~Lars Polsterer.
\newblock Photometric redshift estimation via deep learning. generalized and
  pre-classification-less, image based, fully probabilistic redshifts.
\newblock \emph{Astronomy \& Astrophysics}, 609:\penalty0 A111, 2018.

\bibitem[Glorot et~al.(2011)Glorot, Bordes, and Bengio]{glorot2011relu}
Xavier Glorot, Antoine Bordes, and Yoshua Bengio.
\newblock Deep sparse rectifier neural networks.
\newblock In \emph{Proceedings of the Fourteenth International Conference on
  Artificial Intelligence and Statistics}, volume~15 of \emph{Proceedings of
  Machine Learning Research}, pages 315--323, Fort Lauderdale, FL, USA, 11--13
  Apr 2011. JMLR Workshop and Conference Proceedings.

\bibitem[Guan(2019)]{guan2019localconf}
Leying Guan.
\newblock Conformal prediction with localization.
\newblock \emph{arXiv preprint arXiv:1908.08558}, 2019.

\bibitem[Gupta et~al.(2020)Gupta, Kuchibhotla, and Ramdas]{Gupta2020}
Chirag Gupta, Arun~K. Kuchibhotla, and Aaditya~K. Ramdas.
\newblock {Nested Conformal Prediction and quantile out-of-bag ensemble
  models}.
\newblock may 2020.
\newblock URL \url{http://arxiv.org/abs/1910.10562}.

\bibitem[Izbicki and Lee(2017)]{Izbicki2017}
Rafael Izbicki and Ann~B. Lee.
\newblock {Converting high-dimensional regression to high-dimensional
  conditional density estimation}.
\newblock \emph{Electronic Journal of Statistics}, 11\penalty0 (2):\penalty0
  2800--2831, 2017.
\newblock ISSN 19357524.
\newblock \doi{10.1214/17-EJS1302}.

\bibitem[Izbicki et~al.(2021)Izbicki, Shimizu, and Stern]{Izbicki2021}
Rafael Izbicki, Gilson Shimizu, and Rafael~B. Stern.
\newblock {CD-split and HPD-split: efficient conformal regions in high
  dimensions}.
\newblock \emph{arXiv}, pages 1--31, 2021.
\newblock ISSN 23318422.

\bibitem[Kingma and Ba(2014)]{kingma2014adam}
Diederik~P. Kingma and Jimmy Ba.
\newblock Adam: A method for stochastic optimization.
\newblock \emph{arXiv preprint arXiv:1412.6980}, 2014.

\bibitem[Lei and Wasserman(2014)]{Lei2014}
Jing Lei and Larry Wasserman.
\newblock {Distribution-free prediction bands for non-parametric regression}.
\newblock \emph{Journal of the Royal Statistical Society. Series B: Statistical
  Methodology}, 76\penalty0 (1):\penalty0 71--96, 2014.
\newblock ISSN 13697412.
\newblock \doi{10.1111/rssb.12021}.

\bibitem[Lei et~al.(2013)Lei, Robins, and Wasserman]{Lei2013}
Jing Lei, James Robins, and Larry Wasserman.
\newblock {Distribution-free prediction sets}.
\newblock \emph{Journal of the American Statistical Association}, 108\penalty0
  (501):\penalty0 278--287, 2013.
\newblock ISSN 01621459.
\newblock \doi{10.1080/01621459.2012.751873}.

\bibitem[Rowe et~al.(2015)Rowe, Jarvis, Mandelbaum, Bernstein, Bosch, Simet,
  Meyers, Kacprzak, Nakajima, Zuntz, et~al.]{rowe2015galsim}
Barnaby Rowe, Mike Jarvis, Rachel Mandelbaum, Gary~M. Bernstein, James Bosch,
  Melanie Simet, Joshua~E. Meyers, Tomasz Kacprzak, Reiko Nakajima, Joe Zuntz,
  et~al.
\newblock {GALSIM}: The modular galaxy image simulation toolkit.
\newblock \emph{Astronomy and Computing}, 10:\penalty0 121--150, 2015.

\bibitem[Srivastava et~al.(2014)Srivastava, Hinton, Krizhevsky, Sutskever, and
  Salakhutdinov]{srivastava2014dropout}
Nitish Srivastava, Geoffrey Hinton, Alex Krizhevsky, Ilya Sutskever, and Ruslan
  Salakhutdinov.
\newblock Dropout: A simple way to prevent neural networks from overfitting.
\newblock \emph{Journal of Machine Learning Research}, 15\penalty0
  (56):\penalty0 1929--1958, 2014.
\newblock URL \url{http://jmlr.org/papers/v15/srivastava14a.html}.

\bibitem[Vovk(2013)]{Vovk2013}
Vladimir Vovk.
\newblock {Conditional validity of inductive conformal predictors}.
\newblock \emph{Machine Learning}, 92\penalty0 (2-3):\penalty0 349--376, 2013.
\newblock ISSN 08856125.
\newblock \doi{10.1007/s10994-013-5355-6}.

\bibitem[Vovk et~al.(2005)Vovk, Gammerman, and Shafer]{Vovk2005}
Vladimir Vovk, Alex Gammerman, and Glenn Shafer.
\newblock \emph{{Algorithmic Learning in a Random World}}.
\newblock Springer Science {\&} Business Media, 2005.
\newblock ISBN 0-387-00152-2.

\bibitem[Zhao et~al.(2021)Zhao, Dalmasso, Izbicki, and
  Lee]{zhao2021diagnostics}
David Zhao, Niccol\`{o} Dalmasso, Rafael Izbicki, and Ann~B. Lee.
\newblock {Diagnostics for conditional density models and Bayesian inference
  algorithms}.
\newblock In \emph{Proceedings of the 37th Conference on Uncertainty in
  Artificial Intelligence (UAI)}, volume 125 of \emph{Proceedings of Machine
  Learning Research}. PMLR, 26--29 Jul 2021.

\end{thebibliography}
